\documentclass[conference]{IEEEtran}
\IEEEoverridecommandlockouts
\usepackage{cite}
\usepackage{amsmath,amssymb,amsfonts}
\usepackage{algorithmic}
\usepackage{graphicx}
\usepackage{textcomp}
\usepackage{xcolor}

\usepackage{booktabs}
\usepackage{url}
\usepackage[hidelinks]{hyperref}
\hypersetup{
    colorlinks=true,
    linkcolor=blue,
    citecolor=black,
    filecolor=magenta,      
    urlcolor=cyan,
    pdftitle={Overleaf Example},
    pdfpagemode=FullScreen,
    }

\usepackage{makecell}

\usepackage{xcolor}
\usepackage{color, colortbl}
\definecolor{Gray}{gray}{0.9}
\definecolor{LightCyan}{rgb}{0.9,1,1}
\definecolor{Color2}{rgb}{1,1,0.8}

\usepackage{balance}
\usepackage{amsmath,amssymb,amsfonts}

\def\BibTeX{{\rm B\kern-.05em{\sc i\kern-.025em b}\kern-.08em
    T\kern-.1667em\lower.7ex\hbox{E}\kern-.125emX}}
\begin{document}

\title{Aligning Large Vision-Language Models by Deep Reinforcement Learning and Direct Preference Optimization}

\author{
\IEEEauthorblockN{1\textsuperscript{st} Thanh Thi Nguyen}
\IEEEauthorblockA{\textit{AiLECS Lab, Monash University} \\
% \textit{Monash University}\\
Melbourne, Australia \\
thanh.nguyen9@monash.edu}
\and
\IEEEauthorblockN{2\textsuperscript{nd} Campbell Wilson}
\IEEEauthorblockA{\textit{AiLECS Lab, Monash University} \\
% \textit{Monash University}\\
Melbourne, Australia \\
campbell.wilson@monash.edu}
\and
\IEEEauthorblockN{3\textsuperscript{rd} Janis Dalins}
\IEEEauthorblockA{\textit{AiLECS Lab, Monash University, Australia} \\
\textit{ICMEC Australia, Sydney, Australia}\\
% Sydney, Australia \\
jdalins@icmec.org.au}
% \and
% \and
% \and
}

\maketitle

\begin{abstract}
Large Vision-Language Models (LVLMs) or multimodal large language models represent a significant advancement in artificial intelligence, enabling systems to understand and generate content across both visual and textual modalities. While large-scale pretraining has driven substantial progress, fine-tuning these models for aligning with human values or engaging in specific tasks or behaviors remains a critical challenge. Deep Reinforcement Learning (DRL) and Direct Preference Optimization (DPO) offer promising frameworks for this aligning process. While DRL enables models to optimize actions using reward signals instead of relying solely on supervised preference data, DPO directly aligns the policy with preferences, eliminating the need for an explicit reward model. This overview explores paradigms for fine-tuning LVLMs, highlighting how DRL and DPO techniques can be used to align models with human preferences and values, improve task performance, and enable adaptive multimodal interaction. We categorize key approaches, examine sources of preference data, reward signals, and discuss open challenges such as scalability, sample efficiency, continual learning, generalization, and safety. The goal is to provide a clear understanding of how DRL and DPO contribute to the evolution of robust and human-aligned LVLMs.
\end{abstract}


\section{Introduction}
\label{sec_introduction}

Large Vision-Language Models (LVLMs) have emerged as a powerful class of multimodal systems capable of understanding and generating information across visual and textual domains \cite{liu2023visual,zhuminigpt}. These models underpin a wide range of applications, from image captioning and visual question answering to multimodal dialogue and autonomous agents \cite{fu2024furl}. Their ability to jointly reason over images and text allows LVLMs to perform complex perception and language tasks that were previously challenging for unimodal models. As a result, they are increasingly being adopted in real-world scenarios, including education, healthcare, and human-computer interaction, where nuanced understanding of both visual and linguistic context is essential.

Unlike their LLM counterparts, where the model processes a text prompt and generates a text response, LVLMs take multimodal inputs, typically combining an image with a text prompt, and produce text-based outputs. This enables LVLMs to reason over both visual and linguistic information, allowing for more context-aware and perceptually grounded responses.

\begin{figure*}[!ht]
    \centering    \includegraphics[width=0.70\textwidth]{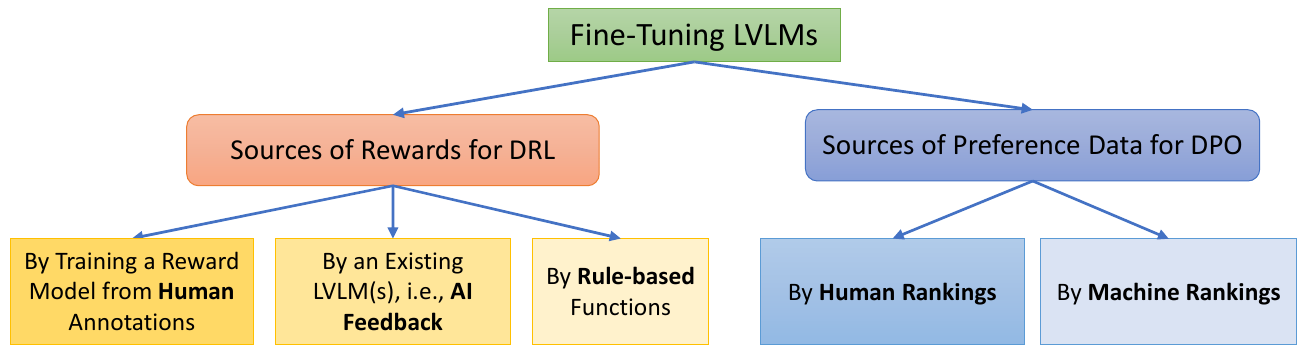}
    \caption{The structure of this survey paper where we categorize preference learning methods for fine-tuning large vision-language models (LVLMs) into Deep Reinforcement Learning (DRL) and Direct Preference Optimization (DPO).}
    \label{fig_structure}
\end{figure*}

Transitioning from LLMs to LVLMs introduces significant complexity due to the inherently multimodal nature of the inputs. While LLMs operate solely on textual data, i.e., taking in a text prompt and producing a corresponding textual output, LVLMs must process and integrate information from multiple modalities, most commonly images paired with text. This added dimension requires the model not only to understand language but also to interpret and align it with visual content, creating a more complex input space and a need for deeper multimodal reasoning.

One of the core challenges in this transition lies in enabling the model to effectively fuse and reason over visual and linguistic signals. LVLMs must develop a coherent understanding of how visual elements relate to textual descriptions and instructions, which can vary significantly in specificity and abstraction. Achieving this requires architectural adaptations, larger and more diverse training datasets, and often more sophisticated alignment techniques to ensure the outputs are both accurate and contextually appropriate. This makes the development of LVLMs more resource-intensive and technically demanding compared to their text-only LLM counterparts.

While pre-training LVLMs on large-scale datasets provides a strong foundation, further training (post-training) is important to refine and align their behavior. Instruction learning and preference learning are complementary approaches used to fine-tune LVLMs, each serving a distinct purpose. Instruction learning focuses on teaching models to follow explicit natural language commands by training on \textit{instruction-response} pairs, i.e., supervised fine-tuning. In contrast, preference learning refines model behavior based on \textit{comparative judgments}, such as rankings or pairwise preferences. While instruction learning helps the model understand and execute tasks, preference learning aligns its outputs with \emph{human preferences and values}, especially in cases where multiple valid responses exist. %Together, these approaches enhance both the functionality and alignment of LVLMs. expectations

%fine-tuning is often essential to align model behavior with specific tasks, user preferences, or real-world constraints. 
This paper focuses on \textit{preference learning} approaches for fine-tuning and aligning LVLMs. As for LLMs, Deep Reinforcement Learning (DRL) has gained traction as a fine-tuning technique that allows LVLMs to learn from feedback signals beyond traditional supervised losses. By optimizing policies with respect to reward functions (whether handcrafted, learned, or derived from human preferences), DRL enables more adaptive, goal-oriented, and interpretable fine-tuning of LVLMs. Direct Preference Optimization (DPO) offers an alternative to DRL for fine-tuning LVLMs using preference data \cite{rafailov2023direct}. Rather than relying on reward functions or policy rollouts, DPO frames preference learning as a classification problem, i.e., directly optimizing the policy to prefer responses aligned with human judgments. This approach simplifies training by avoiding the instability and complexity often associated with DRL algorithms. Moreover, DPO maintains strong performance while offering improved efficiency and more predictable optimization behavior, making it an attractive choice for aligning models with human intent.

This survey explores the intersection of DRL and DPO with vision-language modeling, highlighting key methods, challenges, and recent advances in this rapidly evolving research area. Fig. \ref{fig_structure} provides a graphical overview of these approaches and serves as the organizational framework for this survey. 
%Section \ref{sec_formalism} introduces the formalisms of DRL and DPO for LVLM fine-tuning and summarizes their respective advantages and limitations. Sections \ref{sec_studies_RL} and \ref{sec_studies_DPO} provide in-depth studies on fine-tuning LVLMs using DRL and DPO, respectively. In Section \ref{sec_studies_RL}, we examine the pros and cons of three reward sources for DRL: human feedback, AI feedback, and rule-based rewards. Section \ref{sec_studies_DPO} compares two sources of preference or ranking data for DPO: human- and machine-generated rankings. To support future research, Section \ref{sec_datasets} summarizes available preference datasets. Finally, Section \ref{sec_future_directions} outlines future research directions, followed by concluding remarks in Section \ref{sec_conclusion}. 

\section{Preference Learning for LVLM Fine-Tuning}
\label{sec_formalism}
Preference learning plays a crucial role in fine-tuning LVLMs, enabling them to better align with human values, intent, and expectations. Unlike instruction learning, which relies on explicitly labeled data, preference learning leverages comparative judgments, such as ranking or pairwise comparisons, to guide model updates. This approach is particularly valuable in scenarios where absolute correctness is hard to define, but relative preferences are easier to obtain, such as evaluating image-caption alignment, visual reasoning quality, or stylistic appropriateness. By integrating human or AI-generated preferences into the fine-tuning process, methods like DRL and DPO allow LVLMs to refine their outputs beyond what is achievable with standard training alone, leading to more nuanced, contextually aware, and user-aligned behaviors.

\begin{figure*}[!ht]
    \centering    \includegraphics[width=0.75\textwidth]{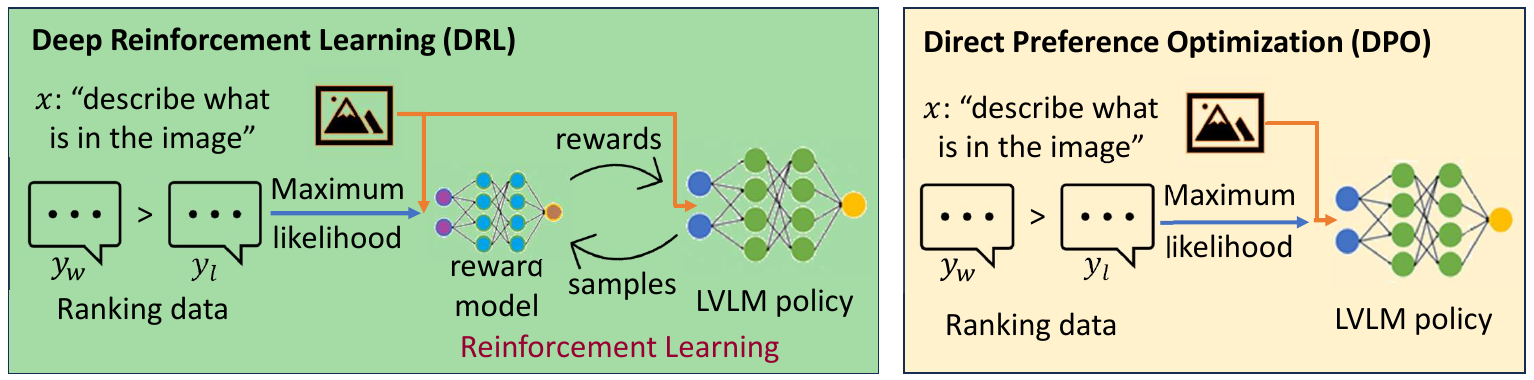}
    \caption{The differences between DRL (left) and DPO (right) approaches for fine-tuning LVLMs (adapted from \protect\cite{rafailov2023direct}). Unlike LLM fine-tuning, images can be used as part of the input during the fine-tuning process of LVLMs.}
    \label{fig_difference_RL_DPO}
\end{figure*}

The difference between DRL and DPO approaches for fine-tuning a LLM or LVLM is depicted in Fig. \ref{fig_difference_RL_DPO}.
DRL-based approaches use preference data to train a separate reward model, which then guides policy optimization through reward signals from that model. %Notably, in DRL, preference data are used exclusively for building this reward model. 
To avoid the time-consuming process of collecting preference data and training a reward model in the DRL approaches, some researchers have explored \textit{rule-based reward} functions as alternatives. In contrast, for DPO, preference data are directly utilized during training, as DPO leverages the model itself to compute preference-aligned gradients, effectively serving as its own implicit reward model.

\subsection{DRL Formalism for LVLM Fine-Tuning}
DRL methods fine-tune LVLMs by formulating the problem as a Markov decision process $(S,A,P,R)$, denoting states, actions, transition probabilities, and rewards, respectively. The state includes both image and text, which serve as input to the model. The state space $S$ therefore is defined as $(I,T_{in})$ where $I$ indicates images and $T_{in}$ represents the input texts. The state $s_t$ includes image, prompt, and past generated tokens, i.e., $s_t=(image, prompt,token_1,token_2,...,token_{t-1})$. The action $a_t$ is choosing the next token $token_t$ from the vocabulary. The action space $A$ therefore includes all possible utterances in the vocabulary and can be denoted as $T_{out}$. Performing the action $a_t$ will transit the model (i.e., state transitions $P$) to the next state $s_{t+1}=(image, prompt,token_1,token_2,...,token_t)$. Designing a reward function $R$ is arguably the most challenging part, whether it is learned via a reward model or crafted using heuristic rules (i.e., rule-based rewards). 
%for timestep $t$

Proximal Policy Optimization (PPO) is one of the popular algorithms used for fine-tuning LLMs or LVLMs \cite{schulman2017proximal}. PPO offers a key advantage over value-based or other policy gradient methods \cite{nguyen2021deep} by balancing ease of implementation with reliable and stable training performance through its clipped objective function, which effectively prevents large, destabilizing policy updates. Recently, DeepSeek introduced a variant of PPO, called Group Relative Policy Optimization (GRPO) \cite{shao2024deepseekmath}. Given the policy model to be optimized $\pi_\theta$ and the reference model $\pi_\text{ref}$, the old policy model ${\pi_\theta}_{old}$ first creates a \textit{group} of completions ${o_1,o_2, ..., o_N}$ for a query $q$. The group rewards ${r_1,r_2, ..., r_N}$ are then calculated for these completions. These rewards are used to compute the advantage $A_i$ for each completion within the group using the formula \cite{zhan2025vision}:
\[
A_i=\frac{r_i-mean(\{r_j\}^N_{j=1})}{std(\{r_j\}^N_{j=1})}
\]
where $N$ is the number of completions in the group. The policy model $\pi_\theta$ is optimized using the GRPO objective function as follows:
\small
\[
\mathcal{L}(\theta)=\frac{1}{N}\sum_{i=1}^{N}\left(\frac{\pi_\theta(o_i|q)}{{\pi_\theta}_{old}(o_i|q)}A_i-\gamma KL(\pi_\theta(o_i|q)|\pi_\text{ref}(o_i|q)) \right)
\]
\normalsize
where $\gamma$ is a hyper-parameter. This objective encourages the model to favor completions with higher advantages within a group, while minimizing deviation from the initial model.

\subsection{DPO Formalism for LVLM Fine-Tuning}
Once preference data are collected, either through human or machine annotations, they can be used to fine-tune LVLMs directly, without the need to explicitly construct a reward model as required in DRL. The DPO approach uses a classification loss function to learn the relative preference between positive (chosen) and negative (rejected) responses. %This is a supervised learning approach by employing two models (policy and reference) and optimize one model towards the better. 
Given the model with parameters denoted as $\theta$, a query as $x$, an image as $I$, positive responses as $y_w$, and negative responses as $y_l$, the DPO objective function is specified as \cite{rafailov2023direct}:
%\small
\fontsize{8.5}{8.5}
%\begin{equation}
%\end{equation}
\[
\mathcal{L}(\theta)=-\mathbb{E}_{(x,y_w,y_l)}\left[\log\sigma\left(\beta\log\frac{\pi_\theta(y_w|x,I)}{\pi_{\text{ref}}(y_w|x,I)}-\beta\frac{\pi_\theta(y_l|x,I)}{\pi_{\text{ref}}(y_l|x,I)}\right)\right]
\]
\normalsize
where $\pi_\theta$ is the policy being optimized, $\pi_\text{ref}$ is the reference policy, $\beta$ is a scaling factor, and $\sigma(.)$ is the sigmoid function. 

In multimodal modeling, DPO operates on paired data consisting of a multimodal input (e.g., an image and a text prompt) along with two responses, where one is marked as preferred. The model is then trained to increase the likelihood of the preferred response over the rejected one. Formally, DPO maximizes the log-likelihood ratio between the chosen and rejected outputs, aligning the model's output distribution with observed human preferences. This makes DPO particularly well-suited for multimodal scenarios where collecting scalar rewards is challenging but preference data are more feasible to obtain. Through this approach, LVLMs can be fine-tuned to generate more helpful, accurate, and aligned outputs across a wide range of multimodal tasks.

\subsection{Merits and Demerits of DRL and DPO for LVLM Fine-Tuning}

Table \ref{tab_rl_vs_dpo} summarizes the advantages and disadvantages of DRL and DPO approaches for fine-tuning LVLMs, highlighting the distinct trade-offs associated with each method. DRL offers high flexibility in reward specification, allowing for complex, multi-signal objectives and the use of pre-existing reward models or rule-based rewards without necessarily requiring preference data. It enables fine-grained behavioural control and has been successfully applied in systems like GPT-4V \cite{GPT4V}, Gemini 1.5 \cite{team2024gemini}, Claude 3 \cite{Claude3Family}, and DeepSeek-R1~\cite{guo2025deepseek}. However, DRL comes with high computational cost, a complex multi-stage pipeline involving reward modelling and policy optimization (e.g., using PPO), and potential training instability due to reward hacking or sensitivity to hyperparameters.

\begin{table}[ht]
\centering
\caption{A comparison between DRL and DPO for fine-tuning LVLMs across different features}
\scriptsize
\begin{tabular}{p{1.3cm}|p{3.2cm}|p{3.0cm}}
\hline
\textbf{Features} & \textbf{Deep Reinforcement Learning (DRL)} & \textbf{Direct Preference Optimization (DPO)} \\
\hline
Dependence on Data & Low (may use existing reward models or rule-based rewards, so not always require preference data) & High (requires explicit preference or ranking data for training) \\
\hline
Complexity & High (multi-stage pipeline with reward modeling and policy optimization) & Low (single-stage supervised learning approach) \\
\hline
\makecell{Training \\Stability} & Can be unstable; sensitive to reward design and hyperparameters & Generally stable; based on supervised training techniques \\
\hline
\makecell{Resource \\Requirements} & High (requires sampling, PPO training, reward model inference) & Low (does not require reward modeling or PPO) \\
\hline
Flexibility & High (supports complex, multi-signal reward objectives) & Moderate (limited to binary preference pairs) \\
\hline
Performance & Strong, allows fine-grained behavioral control & Strong, often matches or exceeds DRL in benchmarks \\
\hline
Best Use Cases & Complex alignment tasks with custom goals & Large-scale preference tuning with efficiency \\
\hline
\end{tabular}
\label{tab_rl_vs_dpo}
\end{table}
% \vspace{-0.6cm}
In contrast, DPO streamlines the alignment process through a stable and efficient supervised learning framework that operates directly on preference data. It is simpler to implement, requires fewer resources, and often matches or exceeds DRL in performance on some alignment benchmarks. DPO has been successful in fine-tuning Llama 3 \cite{grattafiori2024llama} and Qwen2.5-VL \cite{bai2025qwen2}. Nonetheless, DPO is inherently dependent on binary preference or ranking data and offers less flexibility in defining nuanced reward structures, which can limit its applicability in highly customized or multi-objective alignment scenarios. Given the respective advantages and drawbacks of DRL and DPO, a hybrid approach has been employed to fine-tune the latest iteration of Llama, i.e., Llama 4 \cite{llama4}. 

\section{Studies on Fine-Tuning LVLMs by DRL}
\label{sec_studies_RL}

Most studies in this area rely on the standard PPO algorithm or its variants to fine-tune LVLMs, which are summarized in Table \ref{tab:summaryRL}. Some studies employ standard PPO as the policy optimization method, while others adapt or modify PPO or GRPO to better suit specific use cases and improve upon the baseline performance. A critical component of these methods is reward design. We categorize reward sources into three types: human feedback, AI feedback, and rule-based rewards. \emph{Human feedback} involves collecting human preference or ranking data to train a separate reward model. An example of this approach can be found in \cite{sun2024aligning}. \emph{AI feedback} refers to rewards generated by existing models. 
For example, the work in \cite{yu2023fusing} computes the rewards based on outputs of CLIP-based text encoder and image encoder \cite{radford2021learning}. Specifically, the reward is as follow given an input image $x$ and the corresponding text output $y$:
\[
\alpha \left(\frac{CLIP{\text -I}(x)}{||CLIP{\text -I}(x)||}\cdot\frac{CLIP{\text -T}(y)}{||CLIP{\text -T}(y)||}\right)+\beta
\]
where $CLIP{\text -I}$ and $CLIP{\text -T}$ are CLIP image encoder and text encoder, respectively, and $\alpha=50$ and $\beta=-10$ are normalizing factors that adjust rewards to have zero mean and unit variance during training \cite{yu2023fusing}.

\begin{table}[th]
    \centering
    \caption{A summary of DRL approaches for fine-tuning LVLMs}
    \scriptsize
    \begin{tabular}{p{0.6cm}|p{0.9cm}|p{1.6cm}|p{2.4cm}|p{1.2cm}}
        \toprule
        Studies & DRL Algorithm & Base Models & Reward & Tasks \\
        \midrule
        \cite{sun2024aligning} & Standard PPO & CLIP for vision \& Vicuna for text & \textbf{Human feedback} by training reward model on human preferences & Hallucination reduction \\
        % \makecell{Fuse LLMs with multimodal\;\;\\ prompts \cite{yu2023fusing}} & PPO & \makecell{Cosine similarity between the\\ input image and generated\\ text via CLIP image \& text encoders} \\
\hline
        \cite{yu2023fusing} & PPO-clip & CLIP for vision \& GPT-2 for text & \textbf{AI feedback} based on the CLIP image and text encoders & Captioning, Reasoning \\
        % \makecell{Fine-tune VLM decision\\ maker \cite{zhai2024finetuning}} & PPO & \makecell{Rewards are designed depending\\ on tasks, e.g., 1, 0, -1 upon \\win, draw, loss in Blackjack} & \makecell{Current observation and\\a predesigned prompt} & \makecell{Generate utterance containing \\CoT reasoning and a text action} \\
\hline
        \cite{zhan2025vision} & Extended GRPO & Griffon-G-7B or Qwen2.5-VL-7B & \textbf{Rule-based} reward based on dual format, recall and precision & Object localization, Question answering \\
\hline
        \cite{guo2025improving} & Iterative iRe-VLA & BLIP-2 3B model & \textbf{Rule-based} binary reward, 1 for finishing the task, otherwise 0 & Control tasks \\
\hline
        \cite{zhai2024finetuning} & Standard PPO & Llava-v1.6 for vision \& Mistral-7B for text & \textbf{Rule-based} rewards, e.g., 1, 0, -1 upon win, draw, loss in the Blackjack game & Control \& Complex manipulation tasks \\
        \bottomrule
    \end{tabular}
    \label{tab:summaryRL}
\end{table}

Several studies employ \emph{rule-based rewards}, as demonstrated in \cite{guo2025improving,zhai2024finetuning,zhan2025vision}. Designing rule-based rewards is highly task-dependent. The effectiveness and structure of these rewards often rely on domain-specific knowledge and the particular goals of the task. As a result, what works well for one application may not generalize to others, requiring custom reward functions for each new setting. 
Each reward provision method presents distinct advantages and drawbacks, with its suitability largely determined by the task requirements and available resources.
%Each reward provision method offers unique advantages and limitations, making them more or less suitable depending on the task's nature and the resource constraints.
%making them more or less suitable depending on the task and resource constraints.
%One example of this is in \cite{zhan2025vision} where a reward function is designed based on dual format, recall and precision for the object localization and question answering tasks. On the other hand, \cite{guo2025improving} propose binary reward for robotic control tasks where it is equal to 1 if the robot successfully finished the task, otherwise reward is equal to 0. Similarly, the work in \cite{zhai2024finetuning} assign rewards of, for example, 1, 0, and -1 upon win, draw, and loss, respectively, in the Blackjack game.

\textbf{Human feedback} is considered the gold standard for aligning models with human preferences and values. By collecting ranking or preference data directly from people, this method captures nuanced judgments that are difficult to encode through rules or automate with existing models. It is particularly effective for subjective or open-ended tasks where human intuition and contextual understanding are critical, such as assessing the helpfulness or safety of a response. However, the primary drawback of human feedback lies in its cost and scalability. Gathering high-quality annotations is time-consuming, expensive, and may introduce inconsistency due to annotator bias or disagreement. Additionally, the iterative process of collecting and incorporating human feedback can slow down model development cycles.

\textbf{AI feedback} offers a more scalable and cost-effective alternative. In this approach, rewards are generated using existing models, such as CLIP encoders or other LVLMs, to evaluate the quality of outputs automatically. This method ensures consistent feedback and can be applied at scale without the need for manual annotation. It is especially appealing in settings where high-throughput evaluation is needed. However, AI feedback is limited by the capabilities and biases of the models used to generate it. These models may not perfectly align with human values or capture task-specific subtleties, which can lead to reward signals that diverge from actual user preferences. Furthermore, AI-generated rewards often lack the depth of human judgment, especially in complex or novel domains.

\textbf{Rule-based} rewards, on the other hand, are designed using handcrafted metrics or heuristics tailored to specific tasks. They are simple, transparent, and efficient, making them particularly suitable for environments with clear success criteria, such as robotic control, object localization, or structured games like Blackjack. Since no data collection is required, rule-based methods are easy to deploy and debug. However, their major limitation lies in their lack of generality and expressiveness. %These rewards often fail in tasks that require flexibility, adaptation, or subjective evaluation. 
Crafting appropriate rules for open-ended or poorly defined tasks can be extremely challenging and may result in oversimplified or misleading signals.

\section{Studies on Fine-Tuning LVLMs by DPO}
\label{sec_studies_DPO}

Studies in this area are summarized in Table \ref{tab:summaryDPO}. Some studies modify the original DPO objective function, such as by adding additional components (e.g., in the Preference Optimization in VLLM with AI-Generated Dispreferences (POVID) method or the retrieval-augmented DPO (rDPO) method), to enhance its performance. Other studies adapt the use of DPO itself, for example by incorporating dynamic reward scaling or applying it iteratively. 
%, where preference data are used directly without constructing a separate reward model,
In DPO, the quality and source of the preference or ranking response data become crucial. \textit{Humans or machines} can be employed to rank or score the responses according to an alignment purpose, such as reducing hallucination, increasing trustworthiness, or reducing harmfulness. Scoring schemes differ across studies: some use a numerical scale from 1 to 5 \cite{li2024vlfeedback,zhang2025mm}, others from 1 to 100 \cite{dong2024insight}, while some adopt ordinal rankings, such as response A being preferred over C, which in turn is preferred over B \cite{ouyang2022training}.

\begin{table*}[htp]
    \centering
    \caption{A summary of approaches based on DPO variants for LVLM fine-tuning}
    \scriptsize
    \begin{tabular}{p{0.6cm}|p{3.9cm}|p{2.0cm}|p{6.0cm}|p{3.5cm}}
        \toprule
        Studies & Base Models & DPO variants & Ranking by Human/Machine & Tasks \\
        \midrule
        \cite{zhang2025mm} & LLaVA-OV-7B, LLaVA-OV-0.5B, InternVL-1B & Dynamic Reward Scaling within DPO & \textbf{Human ranking} by over 50 annotators, supported by 8 multimodal research experts & Conversation, OCR, Math, Hallucination, Video Understanding \\
        % \makecell{Fuse LLMs with multimodal\;\;\\ prompts \cite{yu2023fusing}} & PPO & \makecell{Cosine similarity between the\\ input image and generated\\ text via CLIP image \& text encoders} \\
\hline
        \cite{yu2024rlhf} & BEiT-3 for vision, Vicuna 13B for text & Dense DPO & \textbf{Human ranking} by fine-grained segment-level corrections & Hallucination Reduction \\
        % \makecell{Fine-tune VLM decision\\ maker \cite{zhai2024finetuning}} & PPO & \makecell{Rewards are designed depending\\ on tasks, e.g., 1, 0, -1 upon \\win, draw, loss in Blackjack} & \makecell{Current observation and\\a predesigned prompt} & \makecell{Generate utterance containing \\CoT reasoning and a text action} \\
\hline
        \cite{yu2024rlaif} & LLaVA 1.5, OmniLMM & Iterative DPO & \textbf{Machine ranking} using Nous-Hermes-2-Yi-34B & Hallucination Reduction\\
\hline
        \cite{wang2024enhancing} & InternVL2-8B, InternVL2-76B & Mixed Preference Optimization & \textbf{Machine ranking} using InternVL2 models & Reasoning, VQA, Hallucination Reduction \\
\hline
        \cite{dong2024insight} & LLaVA-NeXT-LLaMA3, Qwen-2.5-7B & Iterative DPO & \textbf{Machine ranking} by Qwen2 to filter out incorrect answers & Visual Reasoning \\
\hline
\cite{zhou2024aligning} & LLaVA-1.5 7B & POVID & \textbf{Machine ranking} by GPT-4V to hallucinate the original answers to construct dispreferred responses & Hallucination Reduction, Perception,
Cognition, and Reasoning \\ 
\hline
        \cite{xing2025re} & LLaVA-v1.5-7B and LLaVA-v1.6-Mistral-7B & rDPO (retrieval-augmented DPO)	& \textbf{Machine ranking} by GPT-4o mini to mask responses as rejected ones & Hallucination Reduction, VQA \\
\hline
\cite{wang2024enhancing1} & LLaVA-1.5 and VILA \cite{lin2024vila} & Standard DPO & \textbf{Machine ranking} by the LVLM itself through a well-designed critic prompt to assess responses & Hallucination Reduction, Enhance Comprehension Capabilities \\
        \bottomrule
    \end{tabular}
    \label{tab:summaryDPO}
\end{table*}

On the one hand, \textbf{human rankings} offer high-quality, nuanced evaluations that align closely with human judgment and intent. This makes it particularly valuable for tasks requiring subjective interpretation, such as reasoning, contextual understanding, or ethical decision-making. However, human annotation is time-consuming and resource-intensive. For instance, the work in \cite{zhang2025mm} involved two months of intensive effort by a team of 58 annotators and domain experts to produce preference data. Similarly, the study in \cite{yu2024rlhf} collected detailed human corrections at the segment level, highlighting the granularity and effort involved in such approaches.

On the other hand, \textbf{machine rankings} leverage existing models to generate preference labels automatically. This approach is much more scalable and cost-effective, enabling large-scale training without extensive human involvement. Studies such as \cite{dong2024insight}, \cite{wang2024enhancing}, \cite{wang2024enhancing1}, \cite{xing2025re}, \cite{yu2024rlaif}, and \cite{zhou2024aligning} have shown that machine feedback can be effective in aligning LVLMs, especially when high-quality base models are used for evaluation. However, this method is limited by the biases and imperfections of the models providing the rankings. It may fail to capture subtleties or misalign with true human preferences, particularly in complex or ambiguous tasks.

\section{Available Preference Datasets}
\label{sec_datasets}

Aligning LVLMs using preference or ranking datasets has emerged as a powerful technique to enhance model performance in terms of factuality, helpfulness, trustworthiness, and safety. 
%These datasets generally consist of annotated comparisons between model-generated responses, often in the context of multimodal inputs like text, images or videos. 
Human or machine-generated annotations provide critical feedback that helps align model outputs with desired behavioral norms. %, such as reducing hallucinations or increasing response safety.
A variety of datasets have been developed to support this fine-tuning process, each with unique methodologies and objectives.

Table \ref{tab_datasets} provides an overview of these key datasets, emphasizing the origin of the annotations, whether generated by humans or machines, and outlining the specific annotation methodologies used. We include a \textcolor{cyan}{hyperlink} beneath each dataset name to facilitate access to the corresponding dataset. 

Major technology companies such as OpenAI, Meta AI, and Google typically incur substantial costs to acquire large, high-quality annotated datasets. These datasets are instrumental in achieving the state-of-the-art performance seen in models like ChatGPT, Llama, and Gemini. However, such datasets are rarely made publicly available. As a result, smaller institutions and independent researchers often face significant barriers in developing competitive LLMs or LVLMs. The datasets summarized in Table \ref{tab_datasets} thus provide a robust foundation for improving the alignment and reliability of multimodal models, especially for resource-constrained institutions. %researchers.% and institutions.

% \vspace{-0.2cm}

\begin{table*}[htp]
    \centering
    \caption{Available preference or ranking datasets}
     % that can be utilized for training reward models in DRL-based LVLM fine-tuning or directly applied in the DPO-based fine-tuning methods
    \scriptsize
    \begin{tabular}{p{1.9cm}|p{0.9cm}|p{8.4cm}|p{2.0cm}|p{2.7cm}}
        \toprule
        Datasets & By & Annotation Details & Samples & Domains \\
        %Annotators
\midrule
        \href{https://huggingface.co/datasets/zhiqings/LLaVA-Human-Preference-10K}{Preference-10K} \cite{sun2024aligning} & Human & 28 Amazon Turker annotators are asked to compare two responses generated by the LLaVA 7B base model and pinpoint the less hallucinated or more helpful one. & 10k human preference paired outputs & Reduce hallucination \\
\hline
        \href{https://huggingface.co/datasets/openbmb/RLHF-V-Dataset}{RLHF-V-Dataset} \cite{yu2024rlhf} & Human & Responses are obtained from Muffin and humans are asked to directly correct the hallucinated segments, resulting in a factually optimal output, yielding a segment level incremental preference pair. & 5,733 preference pairs & Enhance trustworthiness; Reduce hallucination \\
\hline
        \href{https://huggingface.co/datasets/yifanzhang114/MM-RLHF}{MM-RLHF} \cite{zhang2025mm} & Human & Over 50 annotators, supported by 8 multimodal research experts with strong English proficiency and academic backgrounds to score, rank, and provide textual explanations for responses by Claude 3.5-Sonnet and Qwen2-VL-72B. & 120k comparison pairs & Three domains: image, video understanding, \& LVLM safety \\
\hline
        \href{https://huggingface.co/datasets/openbmb/RLAIF-V-Dataset}{RLAIF-V-Dataset} \cite{yu2024rlaif} & Machine & Use LLaVA 1.5 as instruction model and LLaVA-NeXT as labeller. Another setting uses OmniLMM as both instructor and labeller. Labellers assign each response with a trustworthiness score and comparison pairs are constructed. & 83,132 preference pairs & Enhance trustworthiness; Reduce hallucination \\
        % (to generate responses)
\hline
        \href{https://huggingface.co/datasets/sqrti/SPA-VL}{SPA-VL} \cite{zhang2024spa} & Machine & Responses are collected from 12 open-source (e.g., QwenVL) and closed-source (e.g., Gemini) LVLMs; and GPT-4V is used to annotate two answers to generate the (rejected, chosen) pairs. & 100,788 samples (image, question, responses) & Enhance safety; covering 6 harmfulness domains. \\
        %chosen \& rejected 
\hline
        \href{https://huggingface.co/datasets/MMInstruction/VLFeedback}{VLFeedback} \cite{li2024vlfeedback} & Machine & Responses are obtained from 12 LVLMs such as GPT-4V, LLaVA-series, InstructBLIP, and Qwen-VL; and GPT-4V is used to assess the response quality, rating the helpfulness, visual faithfulness, and safety. & 82k instructions and comprehensive rationales	& Enhance safety, visual faithfulness, and helpfulness \\
        % from 1 to 5
\hline
    \href{https://huggingface.co/datasets/YangyiYY/LVLM_NLF}{LVLM\_NLF} \cite{chen2024dress} & Machine & Supervised fine-tuning EVA-CLIP-Giant and Vicuna-13b-v1.5 to generate responses. GPT-4 provides feedback consisting of numerical score (1 to 4), reason for scoring, \& overall response quality. & 63k annotated vision-language feedback samples & Enhance helpfulness, honesty; Reduce harmlessness \\
        \bottomrule
        
    \end{tabular}
    \label{tab_datasets}
\end{table*}
% \vspace{-0.9cm}

\section{Future Research Directions}
\label{sec_future_directions}

\subsection{Scalable Human Feedback Integration}
One of the key limitations in current DRL and DPO approaches to fine-tuning and aligning LVLMs is the reliance on expensive and limited human feedback. Collecting preferences, ratings, or demonstrations from humans at scale is not only time-consuming but also subject to inconsistency and bias. A promising future direction involves developing scalable methods for integrating human feedback more efficiently. This could include techniques such as active learning to query the most informative examples \cite{safaei2025active}, semi-supervised learning to expand small feedback datasets \cite{mo2023s}, or leveraging large-scale user interaction data \cite{li2024vlfeedback} in a privacy-preserving way. %Additionally, the use of synthetic feedback \cite{liu2025synth,waite2025rls3}, generated by simulators or smaller teacher models, can help bootstrap learning before involving human evaluators.

\subsection{Sample-Efficient DRL Algorithms}
Vision-language models are typically very large, making them computationally expensive to fine-tune using standard DRL algorithms, which often require thousands or millions of interactions to converge. This raises the need for more sample-efficient DRL techniques that can learn from fewer interactions. Techniques such as off-policy learning \cite{wang2024rl}, experience replay \cite{wang2025vl}, and meta-DRL \cite{zhang2025text} can help improve data efficiency. Additionally, model-based DRL \cite{benechehab2025zeroshot}, where the agent learns a predictive model of the environment or reward function, could allow for faster exploration and optimization. %Combining such methods with architectural innovations and more intelligent exploration strategies can significantly reduce the cost of fine-tuning and aligning LVLMs in real-world scenarios.

% \subsection{Continual and Lifelong Learning}
% Most LVLMs today are trained in a static, offline setting and cannot easily adapt to new tasks or domains without retraining or aligning. DRL provides a mechanism for continual adaptation, allowing models to learn from ongoing experience. Future work should explore lifelong DRL techniques for LVLMs that can update their policies incrementally without forgetting previous knowledge, a major challenge known as catastrophic forgetting \cite{10.1145/3735633}. This could be addressed using regularization techniques and memory-based architectures \cite{yu2024boosting}, or modular policies \cite{huang2025lifelong} that separate long-term skills from short-term adaptations. Achieving continual learning would make LVLMs more robust and versatile in real-world applications where tasks and environments are constantly evolving.

\subsection{Generalization and Robustness}
A critical goal for fine-tuning LVLMs is strong generalization to unseen tasks, inputs, or environments. However, models optimized for specific reward functions or preference data can overfit to training conditions and perform poorly in the wild. Enhancing generalization and robustness involves strategies such as domain randomization, adversarial training, or curriculum learning, where tasks are presented in a structured progression of difficulty \cite{deng2025boosting}. Additionally, evaluating models in diverse, open-ended settings is essential to understand failure modes and limitations \cite{ging2024openended}. %Future research should aim to quantify and improve the transferability of skills learned via DRL or DPO, especially in safety-critical applications.
% saha2024exploring
\subsection{Better Reward Modeling}
Designing effective reward functions remains a major challenge in DRL for LVLMs, particularly when dealing with complex, multimodal objectives such as relevance, creativity, and safety. Handcrafted rewards are often crisp and insufficiently aligned with human judgment, while learned reward models can suffer from generalization issues. Future research should focus on training robust, interpretable reward models that can capture subtle aspects of multimodal behavior. One promising direction is preference-based reward learning \cite{liu2025vlp}, where models learn to rank outputs based on pairwise comparisons rather than absolute scores. %Another is adversarial reward modeling \cite{bukharin2025adversarial}, in which discriminators learn to distinguish human-like outputs from model-generated ones, thus guiding learning toward more human-aligned behaviors.

\subsection{Ethics and Safety}
As LVLMs become more integrated into decision-making and content-generation pipelines, ethical considerations and alignment with human preferences become increasingly important. Future directions include building reward models that reflect ethical norms, avoiding bias, and ensuring transparency in decision-making processes. \emph{Human-in-the-loop} DRL \cite{nguyen2019multi}, where users provide real-time feedback on model behavior, can be particularly effective for this. Safety mechanisms, such as reward penalties for unsafe actions, uncertainty-aware exploration, or inference-time alignment framework \cite{ding2025eta} must also be incorporated to prevent harmful outputs.% in unpredictable scenarios.
% yu2024few \cite{ding2025eta} 
% DRL and DPO offer tools for shaping model behavior through reward signals or supervised preference data, making them natural candidates for value alignment. 

\section{Conclusion}
\label{sec_conclusion}

Aligning LVLMs using DRL and DPO represents a promising frontier in multimodal AI, enabling models to move beyond static supervision toward dynamic, feedback-driven learning. By leveraging preference data and reward signals, whether human-derived, task-specific, or learned, DRL and DPO allow these systems to better align with desired behaviors, adapt to new environments, and perform complex reasoning across vision and language. In DRL approach, each reward provision method presents trade-offs between alignment quality, scalability, and task specificity. Human feedback offers the highest fidelity but at significant cost; AI feedback scales well but is limited by model quality; and rule-based methods are efficient yet constrained in scope. In DPO approaches, sources and quality of preference or ranking data are important. Human ranking ensures fidelity to human values but is labor-intensive, while AI ranking offers scalability at the cost of potential misalignment. The choice between them often involves balancing annotation quality with efficiency and scalability.

Despite its potential, these fine-tuning approaches face notable challenges including sample inefficiency, reward design, and safety concerns. Continued research is needed to develop scalable feedback mechanisms, robust reward models, and more generalizable DRL and DPO algorithms. As LVLMs increasingly interact with the real world, integrating DRL and DPO will be key to building systems that are not only more capable and adaptive but also aligned with human values and expectations. This survey outlines foundational ideas and points toward future innovations that will shape the next generation of vision-language intelligence.

\bibliographystyle{IEEEtran}
\bibliography{ijcai25}
\end{document}